\documentclass[letterpaper, 10 pt, conference]{ieeeconf}  

\IEEEoverridecommandlockouts                              

\usepackage{graphics} 
\usepackage{epsfig} 
\usepackage{mathptmx} 
\usepackage{times} 
\usepackage{amsmath} 
\usepackage{amssymb}  
\usepackage{booktabs}

\usepackage{xspace}
\def\OURS{HapticWorld\xspace}

\newcount\Comments 
\usepackage{array,multirow,graphicx}
\usepackage{cuted}
\usepackage{capt-of}
\usepackage{color}
\usepackage{makecell}
\usepackage{wrapfig}
\usepackage{url}
\definecolor{darkgreen}{rgb}{0,0.5,0}
\definecolor{purple}{rgb}{1,0,1}
\newcommand{\kibitz}[2]{\ifnum\Comments=1\textcolor{#1}{#2}\fi}

\title{\LARGE \bf
HapticWorld: an Interactive World Simulator \\ with Real-time Torque Feedback
}

\author{Shaoting Peng$^{1, 2}$, Litian Liang$^{1}$, Yixuan Wang$^{3}$,\\ Ming Yang$^{1}$, Katherine Driggs-Campbell$^{2}$, Mark Cutkosky$^{4}$, James Jingxi Xu$^{1, 4}$
\thanks{$^{1}$Ant Group}%
\thanks{$^{2}$Coordinated Science Laboratory, Electrical and Computer Engineering, University of Illinois Urbana-Champaign}%
\thanks{$^{3}$NVIDIA}%
\thanks{$^{4}$Mechanical Engineering, Stanford University}%
\\
\url{https://haptic-world.github.io/}
\vspace{-80pt}
}

\begin{document}

\maketitle

\begin{strip}
  \centering
  \includegraphics[width=\linewidth]{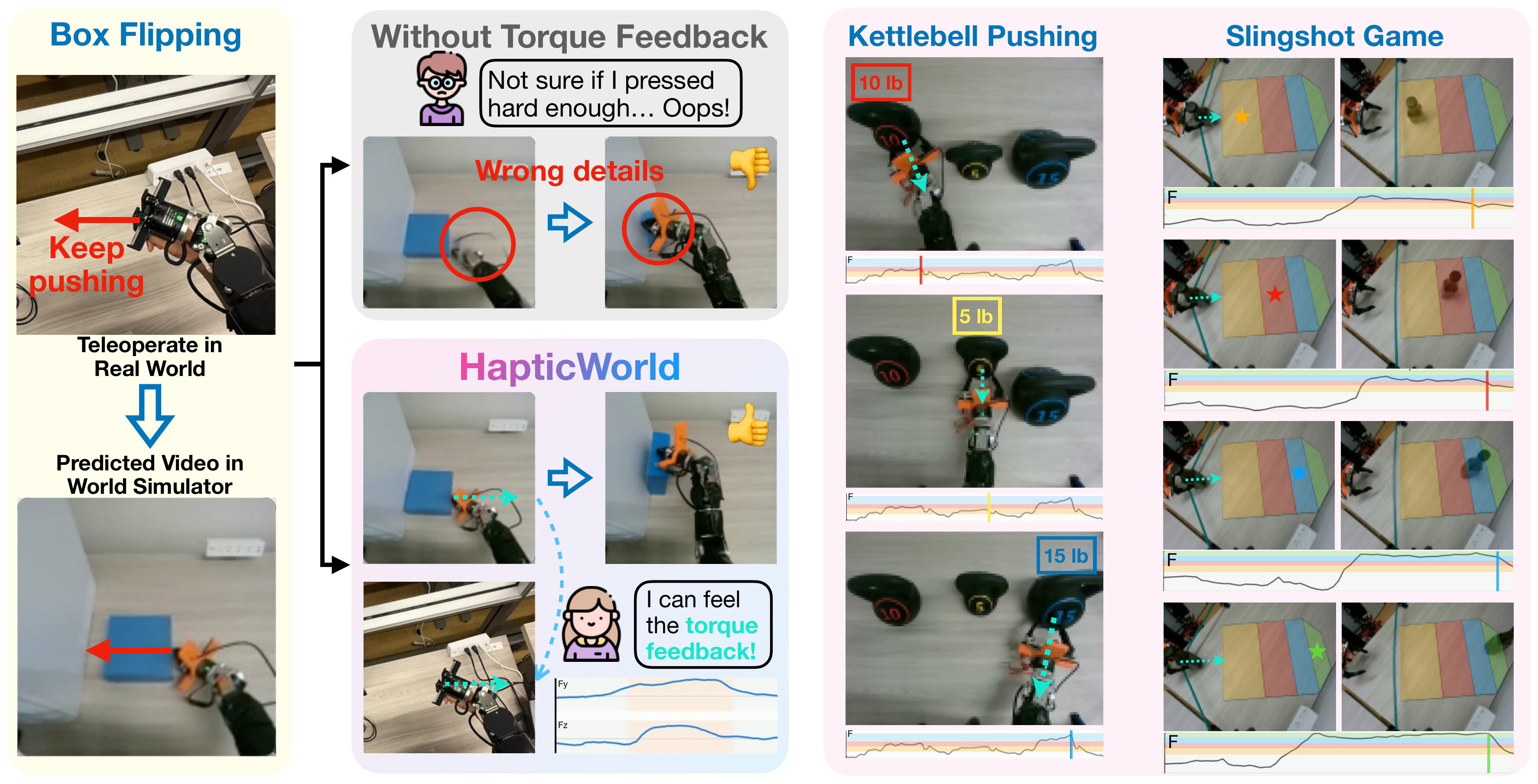}
  \vspace{-20pt}
  \captionof{figure}{\textbf{\OURS closes the haptic loop with a learned world model.}
  Left: the operator teleoperates through a leader arm in real world while the world model predicts the robot keeps pushing the box against the fixture to pivot it upright. Without torque feedback, the operator cannot tell whether enough force is applied, and the rollout drifts out of the distribution the model was trained on and fails the task; with \OURS, the operator feels the predicted torque and flips the box successfully. Right: two game domains illustrate that the rendered torque reflects the physical interaction. Pushing kettlebells of 5, 10, and 15\,lb produces correspondingly larger resistance, and in the slingshot game the draw force and the flight distance grow together, landing the projectile in different scoring zones. Cyan arrows show the force feedback at the end-effector; vertical lines in the force plots mark the frames shown above; and the colored bands give the force range for each behavior: pushing each kettlebell, or landing in each scoring zone.}
  \label{fig:teaser}
  \vspace{-12pt}
\end{strip}
\thispagestyle{empty}
\pagestyle{empty}

\begin{abstract}
Contact-rich manipulation depends on force sensing that is hard to infer from visual signals alone, both for collecting demonstrations and for training policies. Force-annotated data, however, remains hard to obtain at scale: real-robot collection ties every demonstration to physical hardware, physics simulators report contact forces that deviate systematically from real measurements, and learned world simulators, though scalable and realistic, are vision-only, so operators feel nothing during data collection and the data carries no force/torque (F/T) labels. We present \OURS, an interactive world simulator that predicts joint torque together with observations and renders it back to the operator in real time, closing the haptic loop between a human and a learned world model. Across three contact-rich tasks, torque feedback raises data collection throughput by 1.6$\times$ on average. Policies trained on \OURS-generated demonstrations succeed in 54/60 real-world trials, approaching the 56/60 upper bound of real-world data, and far exceeding the 19/60 success rate of the vision-only baseline. Moreover, the success rates measured inside \OURS closely match real-world evaluation, demonstrating that \OURS can serve as a stand-alone F/T-conditioned policy evaluation platform.
\end{abstract}

\section{Introduction}
\label{sec:introduction}

Contact-rich manipulation is governed by forces that visual sensors cannot perceive. Whether a part is jammed or aligned, whether a grasp is about to slip, whether a tool is pressing hard enough to take effect: these states are often visually indistinguishable yet dynamically decisive. Accordingly, a rapidly growing body of work shows that policies observing force or tactile signals substantially outperform vision-only counterparts on contact-rich tasks~\cite{li2022seehearfeel,huang2024vitac,xue2025rdp,yu2025forcevla,xu2022tandem,xu2023tandem3d,xu2025tactile}. Like all imitation-learning policies, however, these force-conditioned policies are data-hungry, and their demonstrations must carry a force channel.

Where can force-annotated demonstrations come from? Hardware-based collection captures real forces~\cite{liu2025factr,liu2024forcemimic,liang2026universal}, but the entire workflow lives in the physical world: the scene must be manually reset after every trial; switching tasks means clearing the old setup, and returning to it later means rebuilding it from scratch; and contact-rich interaction constantly risks damaging the robot or the objects. 
Physics simulation removes the hardware~\cite{todorov2012mujoco,mandlekar2023mimicgen}, but building a simulated scene requires object assets, contact parameters, and per-scene system identification, and even then the rendered observations and dynamics differ from their real counterparts, leaving a sim-to-real gap. For our purpose the problem is sharper still: the contact forces reported by physics engines carry artifacts of relaxed contact models and soft-constraint solvers, and deviate systematically from real measurements~\cite{lelidec2024contact,acosta2022validating}. 

In parallel, a third paradigm has emerged on the vision side: action-conditioned video prediction models, or \emph{world simulators}, that let users collect demonstrations by teleoperating \emph{inside} the model and support reproducible policy evaluation with strong sim-to-real correlation~\cite{guo2025ctrlworld,gemini2025veo}. In particular, the Interactive World Simulator (IWS)~\cite{wang2026iws} needs only a few hundred teleoperated episodes per task, and sustains stable real-time rollouts on a single GPU. Trained directly on real images and real dynamics, world simulators require no scene modeling or parameter tuning; resets are instant and exact, a task's setup lives in the model and can be revisited at any time, and excessive force damages nothing, since no robot or object physically exists. Collection and evaluation thus become convenient and efficient.
 
However, existing world simulators are vision-only, which creates three problems. First, the operator feels nothing: pressing into a virtual object is indistinguishable from moving through free space. With no physics to clamp the commanded motion, the operator's actions may drift out of the training distribution, corrupting rollouts and poisoning the recorded action labels. Second, the generated demonstrations carry no force labels, so force-conditioned policies cannot be trained from them. Third, such policies cannot even be rolled out inside the simulator for evaluation, since the force observations they require do not exist. Recent world models do incorporate force or tactile signals, but as observation channels for autonomous rollouts~\cite{lou2026dreamtac,higuera2026vtwm,huang2026vitacworld}; none closes the loop with a human operator, so the demonstrations they generate inherit no human force feedback.
 
We present \OURS, an interactive world simulator that jointly predicts joint torque alongside visual observations and renders the predicted torque back to the operator in real time; to our knowledge, it is the first system to close the haptic loop between a human operator and a learned world model. \OURS is trained on teleoperated play data with recorded joint torque measurements, and requires no object models, physics engine, or per-scene system identification. We train \OURS on five domains: kettlebell pushing and slingshot shooting for qualitative demonstration (Fig.~\ref{fig:teaser}); microwave door opening, whiteboard wiping, and box pivoting against a fixture for quantitatively evaluating the full pipeline (Fig.~\ref{fig:setup} (b-d)).
 
In summary, our contributions are as follows:
\begin{itemize}
  \item We present \textbf{\OURS}, an interactive world simulator that predicts joint torque jointly with observations. Instead of training a separate torque predictor, \OURS reads torque from the internal features of the dynamics model, encouraging consistency between the predicted torque and sampled visual future through a shared dynamics representation. The predicted torques closely match held-out real measurements: across the three tasks, the overall RMSE is only 0.26--0.38,N$\cdot$m, and the average joint-wise RMSE normalized by each joint's peak torque is 3.5--4.8\%.
  \item \OURS's torque feedback improves data collection efficiency in simulation and the success rate of learned policies in real-world. Collection is faster because the operator can feel contact instead of failing and retrying, raising throughput by 1.6$\times$ on average. More importantly, torque-annotated demonstrations collected inside \OURS are sufficient to train torque-conditioned policies that succeed in the real world: policies trained with \OURS data reach 54/60 success trials across three tasks, approaching the 56/60 upper bound achieved with real-world data, whereas demonstrations collected without torque modality yield only 19/60.
  \item \OURS faithfully evaluate the performance of torque-conditioned policies without a real-robot, which cannot be done with vision-only simulators: success rates measured inside \OURS deviate from real-world evaluation by at most three trials out of twenty.
\end{itemize}
\section{Related Work}
\label{sec:related_work}

\subsection{World Models for Robotic Manipulation}

Action-conditioned video prediction models, often called world models, have been widely applied in robotics. Large-scale video foundation models can be fine-tuned into simulators for robot learning~\cite{agarwal2025cosmos}; some works jointly model video and actions so that the same model also acts as a policy or planner~\cite{li2025uva,ye2026wam,kim2026cosmospolicy}; others learn dynamics in a latent space and plan within it~\cite{zhou2024dinowm,hafner2025dreamer4}; and generated rollouts have been used as additional training data for imitation learning~\cite{jang2025dreamgen,gao2026dreamdojo}. Most relevant to us, recent work uses world models for reproducible policy evaluation without physical robots~\cite{guo2025ctrlworld,gemini2025veo,quevedo2025worldgym}. The Interactive World Simulator~\cite{wang2026iws} provides the visual backbone of \OURS: an action-conditioned consistency model efficient enough for real-time interaction inside the model. However, all of these models are strictly vision-only: none predicts interaction force or torque, and none renders any feedback to the operator.

\subsection{Force/Torque- and Tactile-Aware Policy Learning}

Force/torque (F/T) and tactile sensing consistently improves contact-rich manipulation, from multimodal fusion policies~\cite{li2022seehearfeel,huang2024vitac,ai2024robopack} to reactive and compliance-aware architectures~\cite{xue2025rdp,hou2024acp,he2024foar}, and a fast-growing family of F/T- and tactile-aware VLA models now treats wrench and tactile signals as first-class observations~\cite{yu2025forcevla,huang2025tactilevla,zhao2026fdvla}. These methods are the downstream consumers of our system: they demonstrate that force channels improve policies, yet all presuppose a physical robot with F/T or tactile sensors to collect that data, which is the scarcity our in-simulator collection relieves.

\subsection{Acquiring Force Data: Hardware and Simulation}

Force-annotated demonstrations today mostly come from hardware. Bilateral teleoperation feeds measured or estimated follower torques back to the operator~\cite{liu2025factr,liu2026factr2} or renders tactile signals in augmented reality~\cite{xue2025rdp}; handheld devices capture forces robot-free~\cite{liu2024forcemimic}; and mainstream kinematic interfaces record no force at all~\cite{chi2024umi,wu2023gello,zhao2023aloha}. All build on classical bilateral teleoperation and haptic rendering~\cite{hokayem2006bilateral,hannaford2002passivity}, and all keep a physical robot and scene in the loop, with throughput bound by hardware and no path to policy evaluation. Simulation is the other route: physics engines with tactile simulation and demonstration generation~\cite{todorov2012mujoco,makoviychuk2021isaacgym,si2024difftactile,mandlekar2023mimicgen} produce contact data at scale, but solver contact forces deviate systematically from real measurements, and closing the gap requires per-scene identification. Closer to us, learned dynamics models predict contact outcomes from real interaction data~\cite{calandra2018more,ai2024robopack}, and recent tactile world models jointly predict visual and tactile futures~\cite{lou2026dreamtac,higuera2026vtwm,zheng2026omnivta}; ViTacWorld~\cite{huang2026vitacworld} further uses such a model to generate training rollouts and evaluate policies. In all of these, however, rollouts are produced by the model alone and curated afterwards; none renders its predictions back to a human. To our knowledge, we are the first to close the haptic loop between an operator and a learned world model, using the rendered torque to shape collection itself rather than filtering data after the fact.

\section{Method}
\label{sec:method}

\subsection{System Overview}
\label{sec:method_overview}

\OURS is an interactive world simulator that predicts joint torque together with visual observations and renders both back to a human operator in real time. It is built and used in two phases, involving two kinds of data.

\begin{figure}[t]
  \centering
  \includegraphics[width=\linewidth]{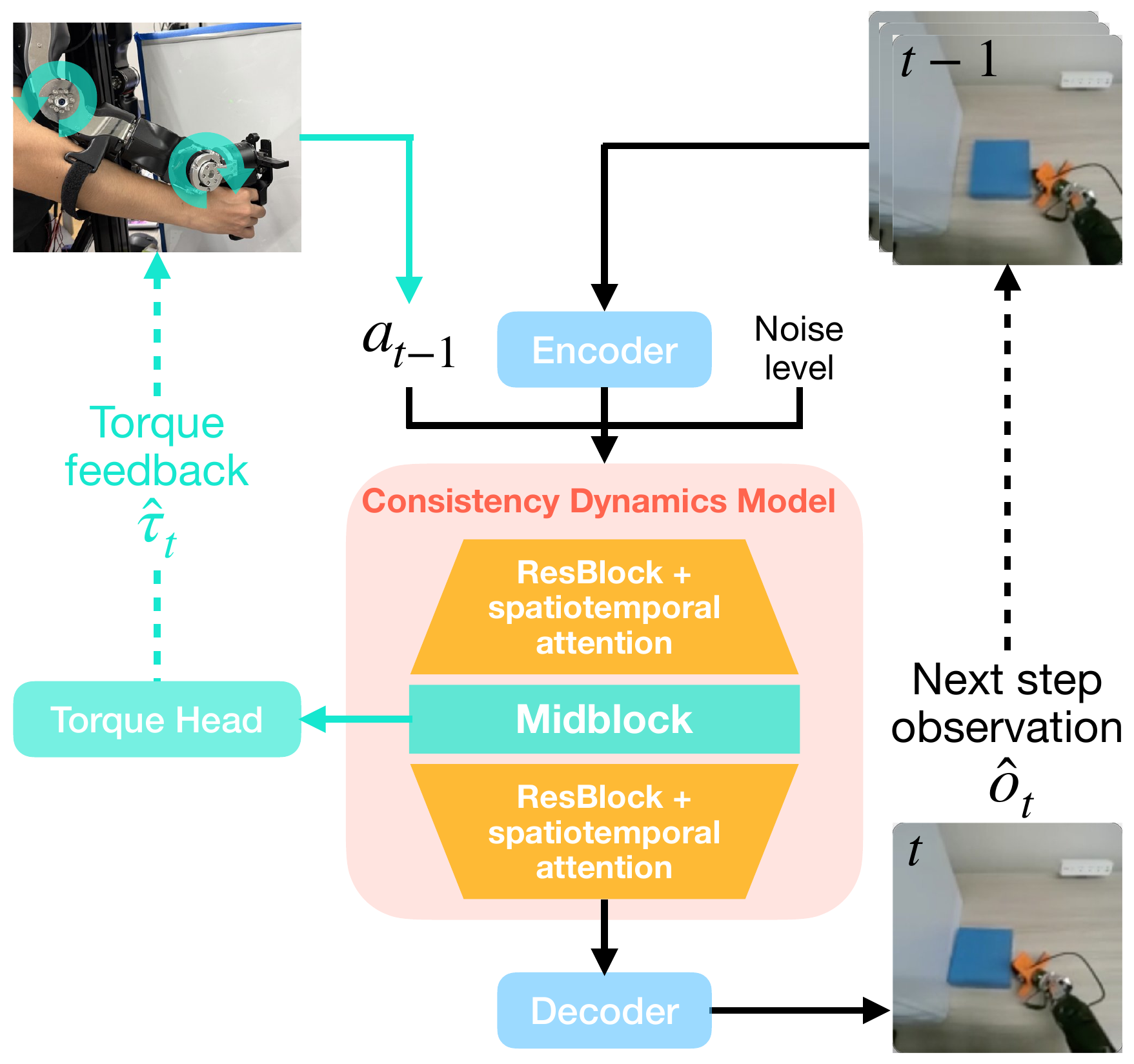}
  \vspace{-20pt}
  \caption{\textbf{\OURS Model Architecture.} At each step, the observation history is encoded into latents and fed to the consistency dynamics model together with the operator's commanded joint positions $a_{t-1}$ and the noise level. The model generates the next latent, which is decoded into the next observation $\hat{o}_t$; the torque head reads the widest intermediate features from the middle of the block stack and maps them to the joint torque $\hat{\tau}_t$, which is rendered back to the leader arm.}
  \label{fig:method}
  \vspace{-10pt}
\end{figure}

In the first phase, we teleoperate the real robot to collect \emph{play data}: task-free interaction in which the operator simply explores the scene, while camera images, commanded joint positions, and measured joint torques are recorded. For example, for the box flipping task play data contains behaviors such as pushing the box in random directions, pivoting it and releasing halfway, or moving the arm without touching the box at all. The play data trains \OURS itself, which couples the visual backbone of the Interactive World Simulator~\cite{wang2026iws} with a torque head, so that every generated frame is accompanied by the joint torque the robot would experience (Fig.~\ref{fig:method}).

In the second phase, the operator teleoperates \emph{inside} the trained model with real-time torque feedback to collect \emph{task demonstration data}: goal-oriented trajectories whose observations and torque labels are both generated by the model. For box flipping, a demonstration is a complete, successful flip. The demonstrations train torque-conditioned policies, which can then be evaluated either inside \OURS or real world. The remaining subsections detail each component.

\subsection{Problem Formulation}
\label{sec:method_problem}

We collect a play dataset $\mathcal{P} = \{\xi^{(1)}, \dots, \xi^{(M)}\}$ by teleoperating the robot with no task objective, where each episode is a synchronized stream of triplets
\begin{equation}
\xi = \big( (o_1, a_1, \tau_1), \; \dots, \; (o_L, a_L, \tau_L) \big).
\end{equation}
Here $o_k \in \mathbb{R}^{3 \times H \times W}$ is the RGB observation at step $k$, $a_k \in \mathbb{R}^{8}$ is the commanded joint position issued by the operator through the leader device, and $\tau_k \in \mathbb{R}^{8}$ is the joint torque measured on the robot, over seven arm joints and the gripper. 
The torque channel can be obtained directly from the robotic arm's motors, without external sensors.

Our goal is a model $f_\Theta$ that, from the past $C$ steps, predicts the next observation together with the torque the robot would experience at that step,
\begin{equation}
(\hat{o}_k, \hat{\tau}_k) = f_\Theta\big( o_{k-C:k-1}, \; a_{k-C:k-1} \big).
\label{eq:problem}
\end{equation}
Here $a_j$ denotes the command issued at step $j$, which drives the transition to step $j{+}1$; $a_{k-1}$ is thus the command the operator has just issued, and it is what the predicted frame and torque respond to.
Two properties of this formulation matter for what follows. First, the two outputs are produced from the same dynamics model rather than separate models, coupling the torque prediction to the representation that generates the visual future seen by the operator. Second, and crucially for torque, the action is the \emph{commanded} joint position rather than the measured one. The two differ when contact occurs. For example, as the eraser presses on the whiteboard, the robot stops at the surface while the commanded position continues below it, and the harder the operator presses, the further below the surface it aims. The observation barely changes as the press deepens, so a model conditioned on measured positions would receive nearly identical inputs across very different torques, and the mapping would not be learnable; the commanded position resolves this ambiguity, because under the arm's position controller the interaction torque is governed largely by the gap between where the robot is told to go and where contact holds it.

\subsection{Action-Conditioned Video Prediction Backbone}
\label{sec:method_model}

The model is trained in two stages. In the first stage, an autoencoder maps each RGB observation $o$ into a compact 2D latent $z = E_\phi(o)$ and reconstructs it with a consistency-model decoder $D_\theta$. In the second stage, the autoencoder is frozen and an action-conditioned latent dynamics model $F_\psi$, also a consistency model, is trained with next-frame supervision: within a window of past latents and actions, full noise is applied to the terminal frame, and $F_\psi$ learns to denoise it conditioned on the action sequence and noise levels. $F_\psi$ is instantiated as a stack of 3D convolutional blocks with FiLM~\cite{perez2018film} action modulation and spatiotemporal attention, and at inference autoregressively predicts latents over a sliding context window, which $D_\theta$ decodes into frames. For the consistency-model training procedure and further architectural details, we follow the Interactive World Simulator~\cite{wang2026iws}.

\subsection{Torque Prediction as a Readout of the Dynamics Model}
\label{sec:method_torque}

Rather than training a separate torque predictor, \OURS reads torque directly from the internal features of $F_\psi$. Let $\Phi_t$ denote the intermediate features at the middle of the block stack of $F_\psi$, where the representation is widest, for frame $t$ during a forward pass of $F_\psi$. These features summarize exactly the information a torque estimator needs: they are conditioned on the commanded actions through FiLM and temporally filtered by causal attention over the context window. A lightweight head $h_\omega$ maps them to per-frame torque,
\begin{equation}
\hat{\tau}_t = h_\omega\!\left(\mathrm{AvgPool}(\Phi_t),\, e_\sigma\right),
\end{equation}
where $\mathrm{AvgPool}$ is spatial average pooling and $e_\sigma$ is the embedding of the frame's noise level, which is provided because the trunk operates on noisy latents and the same physical state yields different features at different noise levels. The head consists of pooling and a small MLP, adding 181K parameters, roughly $0.5\%$ of the full model.

This readout design follows from what the interaction torque actually depends on. The torque at a step is determined by the robot's configuration, the commanded motion, and the state of whatever the robot is in contact with, which is precisely the state the dynamics model must already infer in order to predict how the scene will change. A separate torque network taking raw latents and actions would have to recover that state on its own, effectively duplicating work already performed by the trunk. Reading the trunk instead lets the torque head consume a representation in which the relevant contact information is already exposed, and keeps its parameter count negligible. The arrangement also shapes the trunk itself: because the torque loss backpropagates through $F_\psi$, the features that generate the next frame are additionally trained to carry the contact information needed to predict torque.

Both outputs are trained in a single second-stage run with a joint objective,
\begin{equation}
\mathcal{L} = \mathcal{L}_{dyn} + \lambda\, \mathcal{L}_\tau, \qquad
\mathcal{L}_\tau = \frac{1}{T}\sum_{t=1}^{T} \left\lVert \hat{\tau}_t - \tau_t \right\rVert_2^2 ,
\end{equation}
where $\mathcal{L}_{dyn}$ is the consistency-model denoising loss on the predicted latents, inherited from~\cite{wang2026iws}, torque is supervised densely at every frame of the window. We set $\lambda = 1$.

\subsection{Autoregressive Interaction and Haptic Rendering}
\label{sec:method_render}

At interaction time, \OURS runs autoregressively: at each step the operator's newest command enters the window, $F_\psi$ generates the next latent, and the frozen decoder renders it for display, following the sliding-window inference of~\cite{wang2026iws}. The question specific to \OURS is how torque reaches the operator's hand. First, predicting torque must not slow the loop down. It does not: the torque is read from the same forward pass that generates the latent, so the haptic channel adds no inference cost, and the loop runs at 10\,Hz. Second, the predicted torque is the total joint torque, which includes gravity, while the operator should only feel the interaction. We therefore subtract the configuration-dependent gravity term $g(q)$ computed from a robot model, low-pass filter and clamp the result for safety, and apply it to the motors of the leader device (OpenArm~1) on all eight axes, including the gripper; gripper torque is essential for tasks such as microwave opening, where the operator must feel whether the handle is securely grasped. The haptic rendering can also be disabled while torque prediction keeps running; we use this configuration in our experiments to isolate the effect of feeling the predicted torque during collection, with all other components unchanged.

\begin{figure*}[!t]
  \centering
  \includegraphics[width=\linewidth]{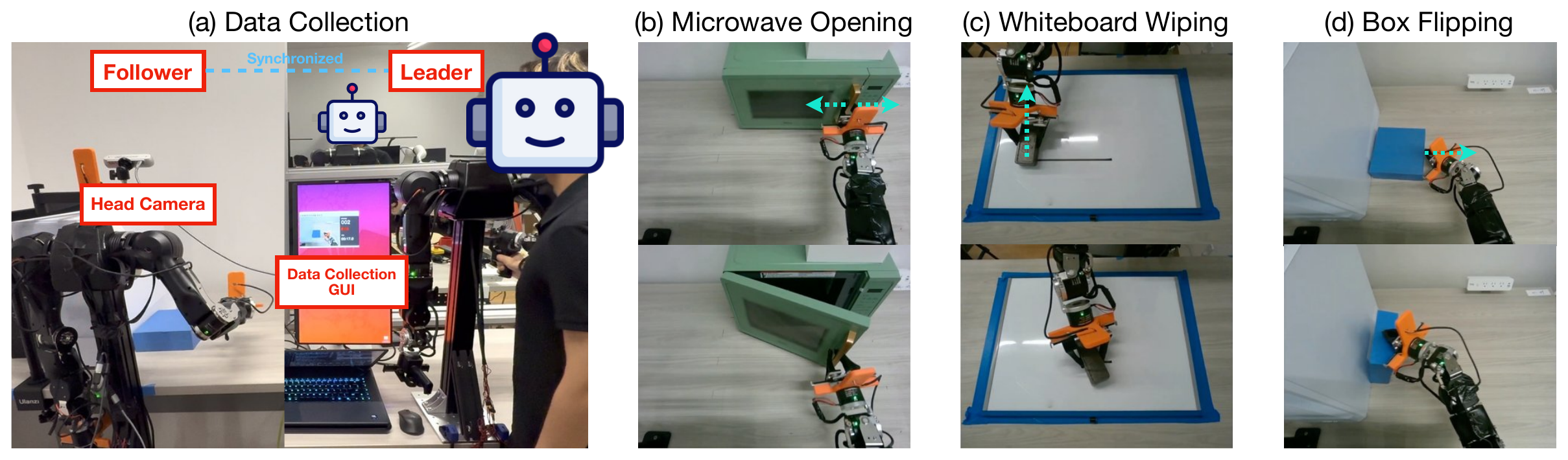}
  \vspace{-20pt}
  \caption{\textbf{Hardware setup and tasks.} (a) Play data is collected by
  teleoperating the follower arm with a synchronized leader arm, recording the head
  camera view, the commanded joint positions, and the measured joint torques.
  (b--d) The three evaluated tasks: open the microwave, wipe the marks on the whiteboard, and flip the box against a fixture. Cyan arrows indicate the equivalent Cartesian force corresponding to the joint-torque feedback.}
  \label{fig:setup}
  \vspace{-13pt}
\end{figure*}



\section{Experiments}
\label{sec:experiments}


\subsection{Experimental Design}
\label{sec:exp_design}

Our experiments are designed to answer four questions, which correspond to the main claims of this work:

\begin{enumerate}
  \item[\textbf{Q1}] Does \OURS predict joint torque accurately? (Sec.~\ref{sec:exp_force})
  \item[\textbf{Q2}] Does haptic feedback improve in-simulator data collection? (Sec.~\ref{sec:exp_collection})
  \item[\textbf{Q3}] Can policies trained on \OURS-generated force-annotated demonstrations transfer directly to real robots? (Sec.~\ref{sec:exp_policy})
  \item[\textbf{Q4}] Can \OURS serve as a stand-alone evaluation platform for force-conditioned policies? (Sec.~\ref{sec:exp_eval})
\end{enumerate}


\textbf{Tasks.} We evaluate on three contact-rich tabletop tasks, illustrated in Fig.~\ref{fig:setup}(b--d):
\begin{enumerate}
    \item \emph{Microwave opening}: Grasping the door handle and pulling it open along the hinge arc, where the grasp must transmit force without slipping. A trial succeeds if the door is opened to at least $45^\circ$. Play data consists of grasping and releasing the handle, and pulling the door to random angles or pushing it back; demonstration data consists of complete openings past $45^\circ$.
    \item \emph{Whiteboard wiping}: Erasing marks from a whiteboard. Whether the eraser is in contact is hard to tell visually, so the operator must rely on torque feedback to judge. A trial succeeds if no marks remain visible. Play data contains wiping in different directions, partial wipes, and wipes that miss the board; demonstration data consists of complete erasures.
    \item \emph{Box flipping}: Pressing a box against a fixture and pivoting it upright, where the normal force must be maintained within the friction cone throughout the rotation to avoid slip. A trial succeeds if the box is brought upright to $90^\circ$. Play data includes random pushes, halfway pivots that are then released, and contact-free arm motion; demonstration data consists of complete, successful flips.
\end{enumerate}

\textbf{Play data collection for \OURS training.} For each task we collect 500 training and 50 validation play episodes of roughly 20\,s each (Fig.~\ref{fig:setup}(a)). Training \OURS takes approximately two days on eight NVIDIA A100 GPUs.

\textbf{Demonstration data collection for policy training.} After training \OURS, We collect 50 demonstration episodes per configuration inside the world simulator and train an ACT~\cite{zhao2023aloha} policy on each, yielding the configurations in Table~\ref{tab:policy_results}. The \emph{IWS} configuration reproduces the complete prior pipeline end to end: the original vision-only simulator~\cite{wang2026iws} is trained on the same play data using only images and actions, and the image-only demonstrations collected inside it train an image-only policy. The \emph{no-feedback} configuration collects inside \OURS with torque prediction running but torque feedback disabled; the demonstrations still carry predicted torque labels, so comparing against the full system isolates the effect of feeling the torque during collection.

\textbf{Evaluation protocol.} Each policy is trained once and evaluated over 20 trials with randomized initial configurations, both in the real world and inside \OURS. Each rollout is limited to 20s.

\begin{figure}[t]
  \centering
  \includegraphics[width=\linewidth]{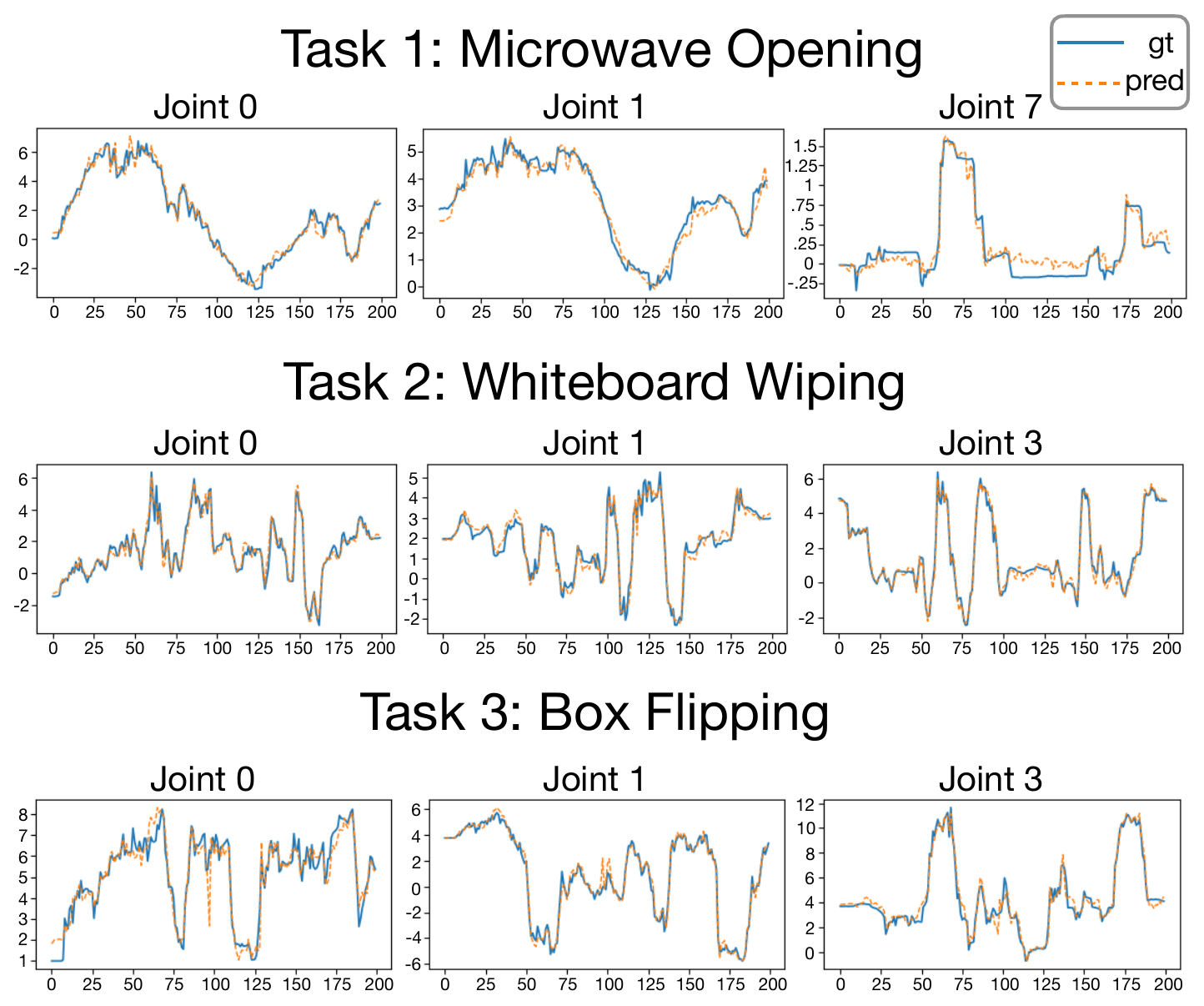}
  \vspace{-20pt}
  \caption{\textbf{Torque prediction on held-out validation episodes.} Predicted (dashed orange line) and measured (solid blue line) torque for three representative joints on each task.}
  \label{fig:torque_pred}
  \vspace{-5pt}
\end{figure}

\begin{figure*}[!t]
  \centering
  \includegraphics[width=\linewidth]{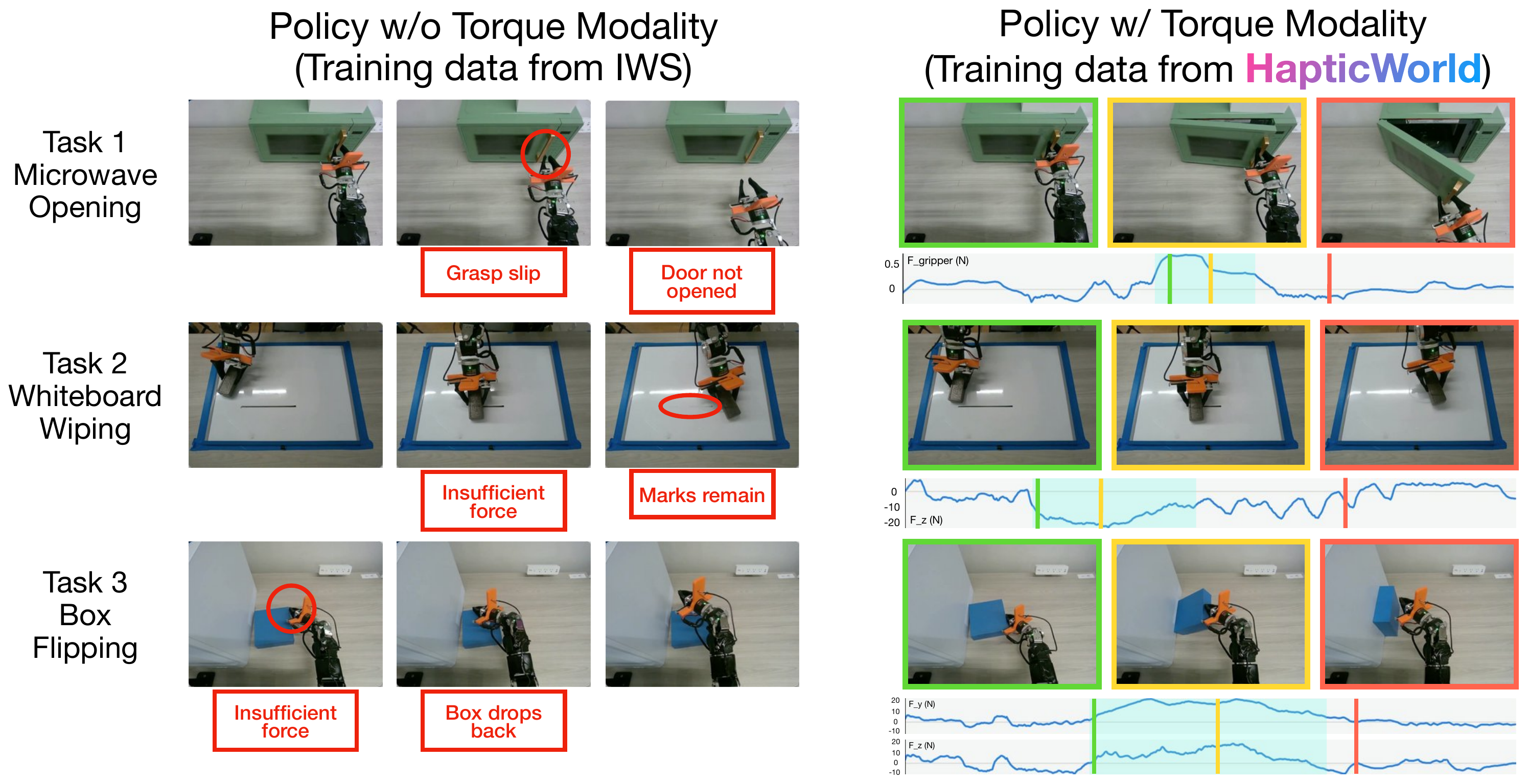}
  \vspace{-20pt}
  \caption{
  \textbf{Real-world rollouts of policies trained without and with torque modality, using data collected from IWS and \OURS, respectively.} 
  Left: failure modes of policies trained without torque: the gripper grasps the microwave handle too weakly and slips, leaving the door unopened (Task 1); the eraser is not pressed hard enough, leaving marks on the whiteboard (Task 2); and the box is not pushed firmly enough and drops back during flipping (Task 3).
  Right: policies trained with \OURS data apply sufficient and consistent force and complete all three tasks. Cartesian force traces derived from the joint torques are shown below each rollout, with the cyan shading marking the phases where torque matters most, and colored vertical lines corresponding to each frame shown above.}
  \label{fig:policy_rollouts}
  \vspace{-8pt}
\end{figure*}

\subsection{\OURS Accurately Predicts Torque (Q1)}
\label{sec:exp_force}


We first verify that the predicted torque is accurate, since every downstream result depends on it. On 50 held-out validation episodes per task, we compare the predicted joint torques against the measured ground truth. The overall torque RMSE is 0.38\,N$\cdot$m on microwave opening, 0.26\,N$\cdot$m on whiteboard wiping, and 0.36\,N$\cdot$m on box flipping; normalizing each joint's RMSE by that joint's peak torque in the same episode, the average relative error is 4.8\%, 3.6\%, and 3.5\%, respectively.
To better illustrate the accuracy, Fig.~\ref{fig:torque_pred} shows predicted and measured traces for the three joints with the largest torque readings on each task: the predictions closely track the ground truth. Full torque visualizations are available on the project website.
\emph{These results answer Q1: \OURS recovers joint torques accurately enough to be rendered to the operator.}

\subsection{\OURS's Torque Feedback Improves Data Collection (Q2)}
\label{sec:exp_collection}

Haptic feedback improves data collection in two ways: it makes in-simulator data collection faster, and improves the data quality.

\textbf{Throughput.} Table~\ref{tab:throughput} reports the number of demonstrations collected in 10 minutes of teleoperation inside \OURS, with and without haptic feedback. Feedback increases throughput on all three tasks, by 1.2$\times$ on microwave opening, 2.4$\times$ on whiteboard wiping, and 1.6$\times$ on box flipping.

\begin{table}[t]
\centering

\caption{\textnormal{\small{Demonstrations collected in 10 minutes of teleoperation inside \OURS.}}}
\label{tab:throughput}
\vspace{-5pt}
\begin{tabular}{lccc}
\toprule
Collection condition & \makecell{Microwave\\Opening} & \makecell{Whiteboard\\Wiping} & \makecell{Box\\Flipping} \\
\midrule
With haptic feedback    & \textbf{26} & \textbf{19} & \textbf{22} \\
Without haptic feedback & 21 & 8  & 14 \\
\midrule
Relative throughput     & 1.2$\times$ & 2.4$\times$ & 1.6$\times$ \\
\bottomrule
\end{tabular}
\vspace{-10pt}
\end{table}

\textbf{Demonstration quality.} Throughput alone does not establish that the retained demonstrations are good. We therefore train identical torque-conditioned policies on 50 demonstrations collected inside \OURS with and without haptic feedback (Table~\ref{tab:policy_results}, second and last columns). Note that, regardless of whether the operator can feel the torque feedback, both policies are trained with the predicted torque modality in \OURS. Removing feedback drops real-world success from 19/20 to 14/20 on microwave opening, from 19/20 to 5/20 on whiteboard wiping, and from 16/20 to 6/20 on box flipping, indicating the importance of \OURS's torque feedback during task data collection.

\textbf{Failure mechanism.} Without torque feedback the operator cannot perceive contact, and applied force falls outside the range covered by the real play data in either direction. Under-application leaves the task incomplete: the door handle is not gripped securely, the eraser does not contact the board consistently, the box is not pressed firmly enough to pivot. Over-application drives the commanded pose into regions the real play data never contains, because during real-world collection physical contact prevents the operator from pressing further; the model has no such constraint, and rollouts degrade. Haptic feedback restores this constraint inside the model, keeping the operator's actions within the distribution the simulator was trained on. Consistent with this account, the benefit is largest on whiteboard wiping, whose two-sided tolerance band is narrowest, and smallest on microwave opening, where over-gripping the handle carries minimal penalty.
\emph{Together these results answer Q2: haptic feedback improves in-simulator data collection by yielding both more demonstrations per unit time and data that train substantially better force-mediated policies.}

\begin{table*}[htbp]
\centering
\caption{\textnormal{\small{Real-world success rates (out of 20 trials) of ACT policies trained on 50 demonstrations per configuration. Policies trained on real-world data serve as a reference upper bound.}}}
\vspace{-5pt}
\begin{tabular}{lcccc} 
\toprule
Data source & IWS~\cite{wang2026iws} & \OURS w/o feedback & Real-world data & \textbf{\OURS (Ours)} \\ \midrule
Task 1 -- Microwave opening & $10/20$ & $14/20$ & $19/20$ & ${19/20}$ \\
Task 2 -- Whiteboard wiping & $1/20$  & $5/20$  & $20/20$ & ${19/20}$ \\
Task 3 -- Box flipping      & $8/20$  & $6/20$  & $17/20$ & ${16/20}$ \\ \bottomrule
\end{tabular}
\label{tab:policy_results}
\end{table*}

\begin{table*}[htbp]
\centering
\caption{\textnormal{\small{Policy success rates (out of 20 trials) for torque-conditioned ACT policies trained on real-world or \OURS demonstration data and evaluated both in the real world and inside \OURS.}}}
\vspace{-5pt}
\begin{tabular}{lcccccc}
\toprule
Training data source & \multicolumn{2}{c}{Real-world data} & \multicolumn{2}{c}{{\OURS data}} \\
\cmidrule(lr){2-3}\cmidrule(lr){4-5}
Evaluation platform & Real-world & \OURS & Real-world & \OURS \\ \midrule
Task 1 -- Microwave opening & $19/20$ & $19/20$ & ${19/20}$ & ${19/20}$ \\
Task 2 -- Whiteboard wiping & $20/20$ & $20/20$ & ${19/20}$ & ${20/20}$ \\
Task 3 -- Box flipping      & $17/20$ & $14/20$ & ${16/20}$ & ${14/20}$ \\ \bottomrule
\end{tabular}
\vspace{-10pt}
\label{tab:eval_results}
\end{table*}

\subsection{\OURS Trains Policies that Work on Real Robots (Q3)}
\label{sec:exp_policy}

Table~\ref{tab:policy_results} compares policies trained on \OURS-generated demonstrations against policies trained on real-world demonstrations, with and without the torque channel.

\textbf{\OURS outperforms the vision-only pipeline.} The complete prior pipeline, collecting inside the vision-only simulator and training an image-only policy, reaches 10/20 on microwave opening, 1/20 on whiteboard wiping, and 8/20 on box flipping, versus 19/20, 19/20, and 16/20 for \OURS (Table~\ref{tab:policy_results}, first and last columns). All three tasks require torque information that neither the visual observation nor the vision-only collection process provides. Fig.~\ref{fig:policy_rollouts} shows the characteristic failures of the image-only policies alongside successful \OURS rollouts.

\textbf{\OURS generated demonstrations match real ones.} Policies trained purely on \OURS demonstrations reach 19/20, 19/20, and 16/20, compared with 19/20, 20/20, and 17/20 for policies trained on real-world demonstrations (third and last columns of Table~\ref{tab:policy_results}), a difference of at most one trial on every task.
\emph{These results answer Q3: \OURS supplies torque data at a quality comparable to real-world collection, so that policies trained with it can be successfully deployed in real world.}

\subsection{\OURS Serves as a Stand-alone Evaluation Platform (Q4)}
\label{sec:exp_eval}

Table~\ref{tab:eval_results} compares each policy's success rate in the real world against its success rate evaluated inside \OURS. Across all three tasks and all evaluated configurations, in-simulator evaluation preserves the ordering of policies and closely tracks real-world success: force-conditioned policies trained on \OURS and on real data score identically in simulation on all three tasks, matching their near-identical real-world performance. Notably, this evaluation is only possible because \OURS predicts torque: a force-conditioned policy can not be rolled out inside a vision-only simulator since the observations it requires do not exist.
\emph{These results answer Q4: \OURS can evaluate the performance of force-conditioned policies faithfully without the real robot and scene, while vision-only simulator cannot.}

\subsection{Additional Tasks with Qualitative Results}
\label{sec:exp_qualitative}

Beyond the three evaluated tasks, we train \OURS on two further domains to illustrate the range of contact behaviors the model captures. In \emph{kettlebell pushing}, three visually distinguishable kettlebells of 5, 10, and 15\,lb sit on the table; pushing each produces a markedly different resistance, giving a direct check that the rendered torque reflects the object being manipulated rather than the motion alone. In \emph{slingshot shooting}, the operator draws a slingshot and releases a projectile into scoring zones, with both the rendered draw force and the resulting flight distance growing with how far the slingshot is pulled. Fig.~\ref{fig:teaser} shows predicted frames alongside the force rendered to the operator in both domains. Please refer to the project webpage for videos and more results.


\section{Conclusion \& Limitation}

We presented \OURS, an interactive world simulator that predicts joint torque jointly with visual observations and renders it back to the operator in real time, closing the haptic loop between a human operator and a learned world model. A lightweight readout of the dynamics network's intermediate features recovers joint torque accurately enough to be felt during teleoperation and consumed by downstream policies. Across three contact-rich tasks, haptic feedback raises collection throughput by 1.6$\times$ on average, and the torque-annotated demonstrations collected inside \OURS are sufficient to train torque-conditioned policies that succeed in the real world, approaching policies trained on real-world data and far exceeding the vision-only pipeline; in-simulator evaluation further preserves real-world policy results for this policy class, which vision-only simulators cannot achieve.

Although promising, several limitations still remain. First, \OURS handles relatively slow and smooth interactions, while fast events such as impacts or sudden slips are neither predicted nor rendered crisply. This stems from both the model's temporal resolution and the 10\,Hz rendering rate, which is below the update rates of the robot. Second, \OURS is trained per task; each new scene requires collecting its own play data and training its own model, and the simulator inherits the coverage of that data. Third, the haptic channel is only as faithful as the visual model it reads from. Since torque is predicted from the same forward pass that generates the frame, errors in the learned dynamics propagate directly into what the operator feels. Addressing fast contact transients, training a single model that serves multiple tasks, and further tightening the physical fidelity of the underlying world model are natural future directions.

\section*{Acknowledgment}
\label{sec:ac}

We thank Houzhu Ding, Jyh-Herng Chow, Kairui Shi, Luqi Wang, Sean Lu, Xinda Qi, Yujun Cai, and Zhixin Sun for insightful discussions and feedback on paper writing, Kairui Shi and Sean Lu for their help in data collection.








\bibliographystyle{IEEEtran}
\bibliography{references.bib}

\end{document}